\documentclass[twoside,11pt]{article}

\usepackage{jmlr2e}
\usepackage{amsfonts}
\usepackage{amsmath}

\newcommand\Tstrut{\rule{0pt}{2.6ex}}         

\ShortHeadings{Multi-task Learning with Weak Class Labels}{Ahmed et al.}
\firstpageno{1}

\begin{document}

\title{Multi-task Learning with Weak Class Labels: Leveraging iEEG to Detect Cortical Lesions in Cryptogenic Epilepsy}

\author{\name Bilal Ahmed \email bahmed01@cs.tufts.edu \\
       \addr Department of Computer Science, Tufts University\\       
       Medford, MA, USA 
       \AND
       \name Thomas Thesen \email thomas.thesen@med.nyu.edu \\
			 \name Karen E. Blackmon \email karen.blackmon@nyumc.edu \\
			 \name Ruben Kuzniecky \email ruben.kuzniecky@nyumc.edu \\
			 \name Orrin Devinsky \email od4@med.nyu.edu \\
       \addr Comprehensive Epilepsy Center, Department of Neurology\\
       School Of Medicine, New York University\\
       New York, NY, USA
			 \AND
       \name Jennifer G. Dy \email jdy@ece.neu.edu\\
			 \addr Department of Electrical and Computer Engineering, Northeastern University\\       
       Boston, MA, USA 
			 \AND
			 \name Carla E. Brodley \email c.brodley@neu.edu\\			 
       \addr College of Computer and Information Science, Northeastern University \\       
       Boston, MA, USA} 

\maketitle

\begin{abstract}
  Multi-task learning (MTL) is useful for domains in which data originates from multiple sources that are individually under-sampled. 
MTL methods are able to learn classification models that have higher performance as compared to learning a single model by aggregating all the data together or learning a separate model for each data source. The performance of these methods relies on label accuracy. 
We address the problem of simultaneously learning multiple classifiers in the MTL framework when the training data has imprecise labels. We assume that there is an additional source of information that provides a score for each instance which reflects the certainty about its label. Modeling this score as being generated by an underlying ranking function, we augment the MTL framework with an added layer of supervision. This results in new MTL methods that are able to learn accurate classifiers while preserving the domain structure provided through the rank information. We apply these methods to the task of detecting abnormal cortical regions in the MRIs of patients suffering from focal epilepsy whose MRI were read as normal by expert neuroradiologists. In addition to the noisy labels provided by the results of surgical resection, we employ the results of an invasive intracranial-EEG exam as an additional source of label information. Our proposed methods are able to successfully detect abnormal regions for all patients in our dataset and achieve a higher performance as compared to baseline methods.
\end{abstract}

\section{Introduction}

\noindent Multi-task learning (MTL) simultaneously learns multiple related prediction tasks which can be represented using a shared common structure \citep{Caruana:1997:MLMTL}. MTL is ideally suited for domains in which the data is collected from different sources, each of which when considered individually does not have enough data to facilitate learning a reliable prediction model. However, because the data are collected from disparate sources, the underlying distribution for each source has its own distinct characteristics. This causes a co-variate shift that negatively impacts the performance of a classifier learned by simply pooling the data together. MTL exploits the ``relatedness'' among tasks by sharing the common information, through joint representation and regularization.

MTL requires accurately annotated training data. However, in many domains there is significant label noise that arises due to human subjectivity, imprecise measurements, etc. Because MTL allows information to be shared among tasks during the learning phase, the impact of label noise can be compounded, undermining model performance. In this paper we formulate an MTL model that learns from imprecise labels, given access to an additional source of information which provides a score for an instance quantifying the confidence associated with its label.  

Our research is motivated by the task of identifying cortical malformations in the MRIs of patients suffering from treatment-resistant epilepsy (TRE) caused by focal cortical dysplasia (FCD) \citep{Bernasconi:FCDReview:NatRev11,Thesen:SBMFCD:PLoS11,Hong:LDAFCD:2014}. The machine learning task is to develop a classifier that can distinguish between normal and abnormal cortical tissue using MRI data. For these patients the only treatment to lead a normal seizure-free life is surgical removal of the affected cortical tissue. However, in patients suffering from FCD which is one of the leading causes of TRE, 45-60\% have a normal MRI \citep{Wang:2013:MRINegRate}. In the absence of a visually detected lesion (\textit{MRI-negative}), the success of surgical resection drops from 66\% to 29\% \citep{Bell:SuccessOutcomeFCD:2009}. From a machine learning perspective detecting FCD lesions for MRI-negative patients has two major challenges: \textit{1) Label noise:} The goal of resective surgery is to remove the entire lesion. If any part of the lesion is left behind, the outcome will not be successful. 
The margin(s) around the lesion is marked in a ``generous'' manner to maximize the likelihood of removing the entire lesion. This introduces false positives in training data.\textit{2) Inter-patient variability:} The morphology of the human brain 
is affected by different demographic factors such as age and gender \citep{Salat:AgingThick:2004}. Therefore, treating the
data from different patients in an identical manner leads to poor classification accuracy.

To address inter-patient variability we treat each patient as a separate classification task. To this end, we use the patient's MRI to isolate the ressected region (positive instances) and extract the same region from an age and gender matched healthy control (negative instances). We then use MTL to learn a common classifier (across all tasks), using the datasets gathered from all the subjects. The common classifier is then used to detect FCD lesions in new patients.

Before undergoing resective brain surgery, all patients are subjected to an invasive intracranial EEG (iEEG) exam. 
A board of certified epileptologists reviews recorded electrical activity to determine the region that is responsible for generating the seizure i.e., the seizure onset zone. To isolate the abnormal region, each electrode is labeled as being part of the seizure onset zone or not. 
However, for MRI-negative patients there is no visible lesion to guide precise electrode implantation, which results in sampling errors 
 in about 40\% of the cases \citep{Hong:LDAFCD:2014}. Nevertheless, iEEG provides valuable information for locating the seizure onset zone. In this work, we augment the MRI labels i.e., the ressected regions for MRI-negative patients with the results of iEEG analysis to mitigate the effects of label noise. 
Using this combined supervision, our proposed approach detects lesions in the resection zones for all MRI-negative patients in our sample, as compared to baseline methods that achieve lower detection rates and/or have higher false positive rates. It should be noted here, that during classification we only have access to patient's MRI data, and the extra supervision from iEEG is used only for training. Thus, our proposed method serves as a pre-surgical patient evaluation tool that detects candidate lesional regions on a patient's MRI data; the candidate lesions are further evaluated using invasive iEEG and video monitoring to locate the final target for resective surgery.

The contributions of this work are: 1) we extend the regularized multitask learning framework \citep{Evgeniou:2004:RegMultiTask} to incorporate auxiliary label information when the training data has weak labels; 2) we model the case when the auxiliary information has similar semantics across tasks, and the case when its semantics differ among tasks; and 3) we cast the task of detecting FCD lesions in MRI-negative patients, as an MTL problem and incorporate the results of their iEEG exams to provide additional supervision to ameliorate the problem of label noise that arises when resection zones are used as ground truth. 

\section{Related Work}
MTL has been successfully applied in various application domains, such as object detection \citep{Torralba:2004:ObjectDet} and conjoint analysis \citep{Lenk:1996:ConjointAnalysis}. A number of different approaches have been taken to develop robust MTL frameworks, including hierarchical Bayes \citep{Thrun96learningto}, regularization \citep{Evgeniou:2004:RegMultiTask}, and Gaussian processes \citep{Wang:2012:SGP}. In this work we extend the regularized MTL framework \citep{Evgeniou:2004:RegMultiTask} to incorporate auxiliary supervision when the training labels are uncertain. We have chosen this framework as it admits a solution similar to support vector learning \citep{Vapnik:1995:NSL}, which along with learning large-margin classifiers allows the use of kernel functions \citep{Scholkopf:2001:LKS}.

Turning now to prior efforts in dealing with label noise, there are two main approaches. The first is to identify the noisy labels \citep{Brodley:99:identifyingmislabeled}, and either discard them or assign them lower weights \citep{Rebbapragada:07:Mitigation}. The second approach assumes that we are provided with scores quantifying the uncertainty of each training label. For example, in learning from probabilistic labels \citep{Smyth:1995:probLabelsVenus,LopezCruz:2013:ProbLabels} the class of each instance is specified by a probability distribution over the possible class labels. Our work falls within this second approach but instead of assuming that the training labels are ``soft'' we assume that we have access to a secondary source of information, providing the means to infer the probability for a particular instance as belonging to either the positive or negative class. In this regard, the work that is closest to ours is Nguyen et al., \citep{Nguyen:2011:LearningAuxProb}, in which the authors assume the availability of an additional source of label information. They model this side information as inducing a pair-wise ranking in the context of learning a binary classifier. The main difference between their approach and our work is that we consider the inclusion of additional label information in the context of MTL rather than for single task learning. We go beyond simply incorporating the additional label information into the regularized MTL framework, by taking two different modeling approaches. In the first approach we assume that the additional source behaves uniformly across tasks, while in the second approach we allow its underlying semantics to be different for different tasks.

\section{MTL with Auxiliary Label Information}
\label{mainSec}
We first provide the details of our notation and review the regularized MTL framework \citep{Evgeniou:2004:RegMultiTask}. 
For clarity and ease of comparison we follow the notation of \citep{Evgeniou:2004:RegMultiTask}.

\noindent \emph{Notation:} We consider that we have data from $T$ related classification tasks given as $(x_{i}^{t},y_{i}^{t})$, where $x_{i}^{t} \in \mathbb{R}^{d}$ and $y_{i}^{t} \in \{-1,1\}$ for all $i \in \{1,2,\ldots,m\}$ and $t \in \{1,2,\ldots,T\}$. All tasks share the same feature space and, without loss of generality we assume that all tasks have an equal number ($m$) of training instances and the underlying data distributions for all tasks are different but related. The goal then is to simultaneously learn $T$ classifiers one per task, such that $f_{t}(x_{i}^{t})=y_{i}^{t}\ ; \ \forall t \in \{1,2,\ldots,T\}$.  

In addition to labeled data we also have a label score $r_{i}^{t}$ for each training instance. This score is considered relative to either the positive or the negative class and represents the degree of ``positive-ness'' or ``negative-ness'' of the instance. For the sake of clarity in the rest of the paper we assume that this score is generated relative to the positive class. This score induces a pairwise ranking of instances for each task: $\left(i,j\right) \in \Pi_{t}\  :\  r_{i}^{t} \geq r_{j}^{t}$. The ranking function $\Pi_{t}$ is adapted from rank-SVM \citep{Joachims:RankSVM}. The score assigned by the ranking function i.e., $r_{i}^{t}$ reflects the degree of belief about an instance belonging to the positive class. 

\subsection{Regularized Multi-task Learning (MTL)}
\label{subsec:RegMTL}
The regularized MTL approach \citep{Evgeniou:2004:RegMultiTask} learns a separate classification function, $f_{t}(x)=w_{t} \cdot x$ for each individual task $t$, defined as: $w_{t} = w_{0} + v_{t}$, where $w_{0} \in \mathbb{R}^d$ is a vector that represents the parameters common to all tasks and $v_{t}\in \mathbb{R}^d$ are the task-specific parameters. If the tasks are highly similar, then the $v_{t}$ are small relative to $w_{0}$, and vice versa. All the classifiers $f_{t}$ can be learned simultaneously by solving the following optimization problem:
\small
\begin{equation}
\label{eqn:mainMTLProblem}
\begin{aligned}
&& \underset{w_{0},v_{t},\xi_{i}^{t}}{\text{min}}
& & & \sum_{t=1}^{T}\sum_{i=1}^{m}\xi_{i}^{t} + 
\frac{\lambda_{1}}{T} \sum_{t=1}^{T} {\left\| v_{t} \right\|}^{2} + 
\lambda_{2} {\left\| w_{0} \right\|}^{2}\\
&& \text{subject to:}\  &&&\\\
&&  \forall i , \ \ \forall t:\ &&&\ y_{i}^{t} \left(w_{0}+v_{t}\right) \cdot x_{i}^{t} \  \geq \ 1-\xi_{i}^{t}\ ,\  \ \xi_{i}^{t} \ \  \geq \ 0 
\end{aligned}
\end{equation}
\normalsize
where, $\lambda_{1}$ and $\lambda_{2}$ are regularization parameters that control the relatedness of the tasks.
The structure of MTL is captured by the kernel function $K_{st}(.)$:
\small
\begin{equation}
\label{eqn:MTLKernelForm}
K_{st}(x_{i}^{t},x_{j}^{s}) = \left( \frac{1}{\mu} + \delta_{st} \right) x_{i}^{t} \cdot x_{j}^{s}
\end{equation}
\normalsize
that arises from the dual of Equation \ref{eqn:mainMTLProblem}. This multi-task kernel couples the inner-product of the instances across tasks based on $\mu=\frac{T \lambda_{2}}{\lambda_{1}}$ which represents the degree of relatedness among the tasks. 
The standard inner product in Equation \ref{eqn:MTLKernelForm} can be replaced with any kernel function \citep{Evgeniou:2004:RegMultiTask}. 

The MTL framework outlined above requires accurate class labels during the learning phase, and the presence of label noise can seriously undermine its performance. We consider the case when the labels are noisy but there is another source of obtaining supplementary label information that can accurately grade the ``positive-ness'' of an instance. This additional source may represent the subjective judgment of a domain expert(s) about a particular instance, based on either the same set of features that are available to the learner or some other view of the data.

In this work we interpret the auxiliary label scores as the output of a pairwise ranking function. 
We can model the behavior of the ranking function as being \emph{globally-consistent} i.e., it does not not vary across tasks. In other words, if we consider the ranking function as representing an expert's judgment, then this assumption would require that his/her evaluation criteria for ascertaining the rank of an instance does not change from one task to another. This is a strong assumption, because most real-life experts will make varying judgments based on the nature of the task at hand. 
We can model this variation by taking a \emph{task-specific} approach that allows each task to have its own ranking function.

From the perspective of MTL, the difference between the two approaches is whether or not the ranking information is shared across tasks. In the \emph{task-specific} approach, we cannot compare the ranks of two instances belonging to different tasks, because the underlying ranking  functions are different. In the \emph{globally-consistent} case, rank information can be shared among tasks without introducing any discrepancies. 
Similar to rank-SVM \citep{Joachims:RankSVM}, we take the rank of each instance as being proportional to its distance from the separating hyperplane:
\begin{equation*}
\label{eqn:semanticsRanking}
\left(i,j\right) \in \Pi_{t}\  :\  w_{t} \cdot x_{i}^{t} \geq w_{t} \cdot x_{j}^{t}
\end{equation*}
where, $\Pi_{t}$ is the ranking function for task $t$. For each pair of instances belonging to task $t$ we augment the original MTL problem (Equation \ref{eqn:mainMTLProblem}) with pairwise rank constraints \citep{Nguyen:2011:LearningAuxProb}. 

\subsubsection{Globally-Consistent Label Ranking (GC)} Here we consider all pairwise ranking functions $\Pi_{t},\ \forall t$ to be identical. In this case we modify the original MTL framework (Equation \ref{eqn:mainMTLProblem}) by adding rank constraints:
\small
\begin{equation}
\label{eqn:primalMTLwithConstraints}
\begin{aligned}
& \underset{w_{0},v_{t},\xi_{i}^{t},\eta_{pq}^{t}}{\text{min}}
&& \frac{1}{2} \sum_{t=1}^{T} {\left\| v_{t} \right\|}^{2} + \frac{\mu}{2} {\left\| w_{0} \right\|}^{2} + C \sum_{t=1}^{T}\sum_{i=1}^{m}\xi_{i}^{t} + C' \sum_{t=1}^{T}\sum_{\left(p,q\right) \in \Pi_{t}}\eta_{pq}^{t} \\
&\text{subject to:} && \\
&\forall i , \ \ \forall t:\ &&\ y_{i}^{t} \left(w_{0}+v_{t}\right) \cdot x_{i}^{t} \  \geq \ 1-\xi_{i}^{t}\ ,\ \  \ \xi_{i}^{t} \ \ \geq \ 0 \\  
&\forall t, \forall \left(p,q\right) \in \Pi_{t}:\ &&\left(w_{0}+v_{t}\right) \cdot \left(x_{p}^{t}-x_{q}^{t}\right) \geq 1-\eta_{pq}^{t}\ ,\ \  \ \eta_{pq}^{t} \ \ \geq \ 0
\end{aligned}
\end{equation}
\normalsize
where $\eta_{pq}^{t}$ are slack variables. 
$C'$ is a positive scalar and represents the relative cost of violating a rank constraint. It is defined as: 
$C'=aC, a\in \mathbb{R}^{+}$. 
The rank constraints can be re-written as:
\begin{equation*}
z_{pq}^{t}\left(w_{0}+v_{t}\right) \cdot \Delta_{pq}^{t} \ \geq \ 1-\eta_{pq}^{t}
\end{equation*}
where, $z_{pq}^{t}=1$ are the labels for each pseudo-example: $\Delta_{pq}^{t}=x_{p}^{t}-x_{q}^{t}$. By augmenting the data of each task with the pseudo-examples $\Delta_{ij}$ and their corresponding labels $z_{ij}=1$ we combine the two sets of constraints to solve a single classification problem. However, the number of pseudo-examples is quadratic in terms of the number of instances in the original dataset for each task. This will cause the number of positive instances to be substantially higher than the number of negative instances, which in the worst case scenario can result in a degenerate solution in which the resulting hyperplanes only respect the rank constraints. The trade-off between preserving the ranking and accurate classification is controlled by the cost parameters $C'$ and $C$, respectively. The cost parameters are analogous to the cost parameter of the traditional support vector machine (SVM) \citep{Scholkopf:2001:LKS}. We set these parameters based on a grid search over a hold-out dataset, which is the standard procedure for training SVMs. The optimization problem in Equation \ref{eqn:primalMTLwithConstraints} can be efficiently solved by formulating its dual using the kernel function from Equation \ref{eqn:MTLKernelForm} \citep{Supplemental:731}\footnote{Please refer to the supplementary material accompanying this paper.}.

It is worth mentioning that no pseudo-examples are created by comparing the ranks of two instances from different tasks. The assumption of global consistency is exploited in the construction of the rank constraints (c.f., Equation \ref{eqn:primalMTLwithConstraints}), in which both $w_{0}$ and $v_{t}$ are required to preserve the ranking. 

\subsubsection{Task-Specific Label Ranking (TS)} To model the peculiarities that may exist in the auxiliary information as we move from one task to another, we can limit the influence that the rank constraints have on the overall solution, by limiting them to affect only the task-specific components. This is formulated similar to Equation \ref{eqn:primalMTLwithConstraints}, with modified rank constraints:
\small
\begin{equation}
\label{eqn:primalMTLwithLocalConstraints}
\begin{aligned}
&&\forall t, \forall \left(p,q\right) \in \Pi_{t}:\ &\ z_{pq}^{t} \  v_{t} \cdot \Delta_{pq}^{t}\geq 1-\eta_{pq}^{t}\\
\end{aligned}
\end{equation}
\normalsize
By not allowing the rank information to directly influence the shared component $w_{0}$, the ranking function $\Pi_{t}$ is no longer coupled across tasks, and can behave differently for different tasks. It should be noted that although the shared weight vector $w_{0}$ is not required to preserve the rankings, it is still indirectly affected by the rank constraints through $v_{t}$ ($\because w_{t}=w_{0}+v_{t}$). This problem can be solved efficiently by formulating its dual \citep{Supplemental:731}, which gives rise to a new kernel function. Let $\tilde{X}^{t} \in \mathbb{R}^{d}$ be the augmented data for task $t$ obtained by combining all the original data instances ($x_{i}^{t}$) and the pseudo-examples ($\Delta_{pq}^{t}$), and let $u_{i}^{t}$ be an indicator variable which is $0$ when the instance is a pseudo-example and $1$ otherwise. Then the new kernel is given by:
%
\small
\begin{equation}
\label{eqn:MTLKernelNew}
K_{st}(\tilde{x}_{k}^{t},\tilde{x}_{l}^{s}) = \left( \frac{u_{k}^{t} u_{l}^{s}}{\mu} + \delta_{st} \right) \tilde{x}_{k}^{t} \cdot \tilde{x}_{l}^{s}
\end{equation}
\normalsize
This is a multitask kernel that does not allow the ranking function $\Pi_{t}$ to directly impact $w_{0}$, restricting the auxiliary label information from being directly shared among tasks.

In both the globally-consistent and task-specific cases, the final optimization takes the form of a standard quadratic program (QP) with box constraints \citep{Boyd:2004:CO} which can be easily solved using any off-the-shelf QP-solver.\footnote{We used the QP solver included in the optimization toolbox for Matlab (http://www.mathworks.com).}

\section{Detecting Cortical Malformations in Epilepsy}
\label{SBM:LesDet}
For patients suffering from treatment-resistant epilepsy (TRE) caused by FCD, early detection and subsequent surgical removal of the lesion area is the only treatment. A comprehensive clinical evaluation for surgery candidates involves a neurological exam, scalp electroencephalography (EEG), neuropsychological exam, and a visual inspection of the patient's MRI. In 45-60\% of such cases, no lesion is detected on the MRI by expert neuroradiologists \citep{Wang:2013:MRINegRate}. Our research focuses on developing machine learning algorithms that can detect these subtle FCD lesions using the patient's MRI data. To this end we use training data that includes patients who underwent surgery and their ressected tissue was histopathologically verified to contain FCD. 

\subsection{Surface-Based Morphometry and Lesion Detection}
Surface based morphometry (SBM) models the cortical surface as a folded two-dimensional surface. It is extracted by delineating the boundary between the gray and white matter using T1-weighted MRI images \citep{Dale:SBA1:1999}. After extraction the cortical surface can be registered to a standard surface also known as a group-atlas. Registration allows for a more precise comparison of individual cortical structures across subjects \citep{Fischl:RegThick:2000}. SBM has been used successfully for analyzing and detecting neurological abnormalities in Schizophrenia \citep{Rimol:Schizo:2012}, Autism \citep{Nordahl:Austism:2007}, and Epilepsy \citep{Thesen:SBMFCD:PLoS11,Hong:LDAFCD:2014}. SBM has been used in conjunction with different machine learning and statistical techniques to identify lesions in MRI-positive FCD patients by classifying each vertex on the cortical surface. These approaches include neural networks \citep{Besson:SBMFCD:MICCAI08}, uni-variate z-score based thresholding \citep{Thesen:SBMFCD:PLoS11}, stratified ensemble of logistic regression classifiers \citep{Ahmed:2015:LabelNoiseMRI}, semi-supervised conditional random fields \citep{Ahmed:ICML:2014} and linear discriminant analysis (LDA) \citep{Hong:LDAFCD:2014}. 

Recall, that one of the major confounding factors in this data is the label noise that arises from using resection zones as ground truth. In our approach to minimize the effects of label noise we do not pre-process the resection zone prior to classifier training based on ad-hoc heuristics \citep{Ahmed:2015:LabelNoiseMRI} or subjective expert opinions \citep{Hong:LDAFCD:2014,Besson:SBMFCD:MICCAI08}. Instead, we augment the resection zones with the results of the iEEG analysis as positive labels for the training data. 

None of the detection schemes address the co-variate shift arising from inter-patient variability in brain morphology and the differences in patterns of seizure onset and intensity. To overcome this co-variate shift
we treat each patient as a separate classification task. To this end, we isolate a patient's resection zone and the same region from an age and gender matched control. Both the lesional and normal images are segmented to obtain super-pixels which are used as data instances. We learn all patient specific classifiers simultaneously using our proposed MTL approaches. In regularized MTL framework (c.f. section-\ref{subsec:RegMTL}), the trained model is tested on previously left-out data from the same tasks that generated the training data. However, in our case test data comes from new patients that were not part of the training data. For classifying data from out-of-sample tasks, we discard task-specific parameter vectors $v_{t}$ and use the mean component $w_{0}$ \citep{VanEsbroeck:2016:throwVT}.

\subsection{Data Description and Pre-processing} 
The dataset consists of 18 MRI-negative FCD patients collected over a three year period from New York University's comprehensive epilepsy center which is a leading level-4 epilepsy treatment center. All 18 patients underwent successful resective surgery and histopathological examination showed evidence of FCD. This may seem a small set but note, that only a few MRI-negative patients proceed to surgery, and out of those only a third have successful outcomes \citep{Bell:SuccessOutcomeFCD:2009}. The controls were matched from a cohort of 115 neurotypical controls. All patients and controls were scanned using the same imaging protocol \citep{Supplemental:731}.

Our training and test data comprises of image patches taken from the resected regions of the patients, and corresponding regions from matched controls. We focus only on the resected regions because all the patients included in our experiments were seizure-free after surgery, which shows that their resected regions contained the primary FCD lesion(s). Furthermore, for new MRI-negative patients, \textit{seizure-semiology} i.e., signs and symptoms of a seizure provides credible but crude estimation about the location of the epileptogenic zone \citep{Rosenow:Semiology:2001,Noachtar:Semilogy:2009,Tufenkjian:Semilogy:2012}. In these cases our methods can be applied to the suspected cortical region.

\emph{Segmentation}: All the patient and control surfaces were registered to an average surface. 
After registration the resected region(s) for each patients and the corresponding region(s) from his/her matched control was isolated, and flattened to a standard 2-d image. We use cortical thickness to represent the intensity values.
It should be noted that even though we use cortical thickness to obtain the super-pixels, a larger set of morphological features (including cortical thickness) is used to characterize each super-pixel.

We use Quickshift \citep{Vedaldi:quickshift:2008} for unsupervised segmentation. 
Quickshift has two parameters: size of the Gaussian kernel ($\sigma_{QS}$) used by a Parzen window density estimator, and the maximum distance ($\delta_{QS}$) between two pixels permitted while remaining part of the same segment. 
All the patients and controls were segmented using the same set of Quickshift parameters ($\sigma_{QS}=8, \delta_{QS}=32$) that were optimized using a parameter estimation set of three patients, which are distinct from the fifteen patients whose results are reported here. After segmentation, each super-pixel is treated as an independent instance. We used first and second-order statistics of cortical thickness, gray-white contrast (GWC), curvature, sulcal depth, Jacobian distance and local gyrification index (LGI) to represent each super-pixel \citep{Thesen:SBMFCD:PLoS11}. Additionally, we also included the average surface area measured on both the pial and white matter surfaces.


\emph{Creating Electrode Maps}: 
The spatial resolution of the iEEG and MRI are not identical: the MRI voxels (and corresponding surface vertices) are smaller than the point spread function that defines the iEEG generators. More importantly, localizing the source for a given electrode is subject to the ill-posed inverse problem \citep{Yuan:iEEG:2012}. These are well-known limitations of iEEG. 
Solving this problem is beyond the scope of this work. However, a sphere with half the diameter of the inter-electrode distance (approximately ten mm), presents a reasonable criterion for matching iEEG and MRI. Therefore, all surface vertices within a radius of five mm (Euclidean distance) from the electrode's location were considered within range of the electrode. We only selected electrodes that were labeled by the experts as either being part of the seizure-onset zone or recorded abnormal electrical activity during seizure onset. 


\section{Results}
\label{results:final}
The lesions of all 18 MRI-negative patients used in our experimental evaluation were located in the temporal region, which is one of the most prevalent localization of FCD in adults \citep{Krsek:temporalPrevalence:2008}. All patients had undergone successful resective surgery and were histopathologically verified to have FCD. We make two sets of comparisons: one designed to evaluate the effectiveness of the MTL framework and the other to evaluate whether including auxiliary supervision in the MTL framework boosts the lesion detection rate over MTL alone.

\begin{table}[t]
\centering
\small
\begin{tabular}{|l||rrrrrr|rrrrrr|}
\hline 
{}&\multicolumn{6}{c|}{\textbf{Recall}}&\multicolumn{6}{c|}{\textbf{False Postive Rate}}\Tstrut\\
\small{Id.}\normalsize&LDA&SLR&SVMR&MTL&GC&TS&LDA&SLR&SVMR&MTL&GC&TS\Tstrut\\
\hline 
{P01}&  {0.07}& {0.04}& {0.48}& {-}& {-}& {0.09}& {-}& {0.05}& {0.22}& {-}& {-}& {0.21}\Tstrut\\
{P02}&  {0.11}& {0.19}& {0.39}& {0.28}& {0.09}& {0.25}& {0.18}& {0.14}& {0.56}& {0.32}& {0.15}& {0.29}\\
{P03}& {0.17}& {-}& {0.52}& {0.22}& {0.02}& {0.11}& {0.07}& {0.03}& {0.26}& {0.08}& {-}& {0.03}\\
{P04}&  {-}& {0.12}& {0.30}& {0.20}& {-}& {0.20}& {0.1}& {0.10}& {0.35}& {0.17}& {-}& {0.13}\\
{P05}&  {0.15}& {0.03}& {0.60}& {0.03}& {-}& {0.03}&  {-}& {0.02}& {0.26}& {0.03}& {-}& {0.03}\\
{P06}&  {0.12}& {-}& {0.62}& {-}& {-}& {0.09}& {-}& {0.05}& {0.39}& {0.04}& {-}& {0.09}\\
{P07}&  {-}& {-}& {0.30}& {-}& {-}& {0.09}& {0.14}& {0.03}& {0.26}& {0.02}& {-}& {-}\\
{P08}&  {0.16}& {0.13}& {0.67}& {0.12}& {0.06}& {0.12}& {0.07}& {0.04}& {0.30}& {0.06}& {0.03}& {0.03}\\
{P09}&  {0.15}& {0.06}& {0.49}& {0.08}& {0.46}& {0.22}& {-}& {-}& {0.48}& {-}& {0.16}& {0.05}\\
{P10}&  {0.04}& {-}& {0.42}& {0.04}& {-}& {0.04}& {0.05}& {0.01}& {0.23}& {0.09}& {-}& {0.13}\\
{P11}&  {0.08}& {-}& {0.37}& {0.07}& {-}& {0.05}& {-}& {0.05}& {0.25}& {0.11}& {-}& {0.15}\\
{P12}&  {-}& {0.05}& {0.57}& {0.05}& {-}& {0.05}& {0.07}& {0.07}& {0.23}& {0.05}& {-}& {0.05}\\
{P13}&  {0.13}& {0.21}& {0.95}& {0.16}& {-}& {0.44}& {0.13}& {0.11}& {0.70}& {0.13}& {-}& {0.14}\\
{P14}&  {-}& {-}& {0.28}& {0.14}& {-}& {0.21}& {0.02}& {0.16}& {0.65}& {0.16}& {-}& {0.15}\\
{P15}&  {0.09}& {0.15}& {0.53}& {0.04}& {-}& {0.1}& {0.15}& {0.05}& {0.39}& {-}& {-}& {0.12}\\
\hline 
{{Mean}}& {0.09}& {0.07}& {0.50}& {0.10}& {0.04}& {0.14}& {0.07}& {0.06}& {0.37}& {0.08}& {0.02}& {0.11}\\
\hline 
\end{tabular}
\caption{Detailed results for MRI-negative subjects. LDA is the Fisher linear discriminant analysis based method \citep{Hong:LDAFCD:2014}, SLR represents the stratified classification scheme \citep{Ahmed:2015:LabelNoiseMRI}, SVMR is a single task SVM with rank constraints \citep{Nguyen:2011:LearningAuxProb}, MTL represents regularized MTL \citep{Evgeniou:2004:RegMultiTask} without rank constraints, GC and TS are the globally-consistent and the task-specific approaches, respectively (`-' represents a value of zero).
\vspace{-0.8cm}}
\label{tableMainResults}
\end{table}

\emph{Baseline Selection}: To show that the regularized MTL framework \citep{Evgeniou:2004:RegMultiTask} can be used effectively to detect FCD lesions, we adapt the LDA based classification scheme \citep{Hong:LDAFCD:2014} as one of the baselines. To this end we use LDA to classify super-pixels using the same set of features as used by our proposed MTL methods. Additionaly, we also compare the performance of our proposed methods to a stratified vertex-based classification approach \citep{Ahmed:2015:LabelNoiseMRI} (SLR). In all of our experiments we have not used any post-processing for any of the proposed methods or baselines.


\emph{Experimental Setup}: A leave-one-patient-out cross-validation strategy was used, in which we left out one patient's data and trained on the remaining patients. Hyperparameters for the proposed methods and baselines were set during each round, using the data of three MRI-negative patients and their matched controls \citep{Supplemental:731} 
 whose iEEG data was not available. We will refer to this set of three patients as the model parameter set (MPS). Data in MPS are distinct from the patients and controls used in our experiments.

\emph{Performance Analysis}: We have developed the proposed methods keeping in view their final use as focus-of-attention tools for neuroradiologists for detecting visually elusive FCD lesions. Therefore, the detection rate i.e., the number of patients whose lesions are correctly detected constitute the main result. For a more detailed performance analysis, we calculate the recall and false positive rate (FPR) based on surface area. Recall represents the percentage of the surface area on the patient's resection zone correctly identified as lesional, and FPR is the percentage of the surface area incorrectly labeled as lesional on the matched control's selected region. Note that the estimates for recall should be considered as \emph{lower bounds} because they are calculated using the noisy resection zones as ground truth. 


Table-\ref{tableMainResults} compares the detection rate (number of patients with positive recall), recall and FPR of the proposed \emph{task-specific} (TS) and the \emph{globally-consistent} (GC) approaches with baseline methods. Among the baselines that do not rely on iEEG data, MTL outperforms both LDA and SLR by correctly detecting the lesions in twelve patients, whereas LDA and SLR detect the lesion in eleven and nine patients, respectively. Not only does MTL achieve a higher detection rate, it also has higher average recall than both LDA and SLR. On the other hand, TS detects the lesion in all fifteen patients and achieves higher average recall as compared to LDA, SLR and MTL. Both TS and SVMR have the same detection rate, but TS clearly outperforms SVMR as far as average FPR is concerned, the average FPR of TS is significantly lower than that of SVMR which is unacceptably high (37\%), when interpreted in the context of its use as a focus-of-attention tool. The lower FPR of TS shows that an MTL based formulation coupled with auxiliary supervision leads to superior detection rate and lower FPR. Turning now to GC, we see that it has the worst performance in terms of detection rate among all the methods. 
The low detection rate of GC when compared to TS, substantiates our assumption that the information obtained from the iEEG analysis has task-specific semantics and the criteria used for determining the seizure-onset zone differ on a patient by patient basis. When this information was shared freely among tasks, the label noise was further compounded leading to low detection rate of the GC method.

\section{Conclusion}
\label{cnclsn}
In this work we addressed the problem of MTL in the presence of uncertain labels, assuming an additional source of supervision, that we modeled as a pairwise ranking function. To this end, we extended the regularized MTL framework \citep{Evgeniou:2004:RegMultiTask} by incorporating additional rank constraints. We modeled the case when there is a single ranking function for all tasks, and the case where each task has its own ranking function. In the latter \textit{task-specific} case we developed a new operator-valued kernel \citep{Evgeniou:2005:LMT} for ensuring that ranks are not directly shared between tasks. In all cases, the model parameters were found by solving a quadratic optimization problem (QP) with box-constraints \citep{Boyd:2004:CO}, which is solvable with any standard QP solver. 
We demonstrated the efficacy of the proposed method on the challenging problem of detecting epileptogenic lesions. 
The task-specific approach detected lesions within the resections of all patients, whereas the baselines either achieved lower detection rates or had an unacceptably high false positive rate. Identifying the abnormal region in cryptogenic epilepsies is based on a confluence of evidence from multiple sources such as MRI, PET, etc. In this regard, our proposed method will have a deeper impact in the application domain by enhancing the sensitivity of the patient evaluation methodology. Furthermore, the proposed methods can be broadly applied to other domains in which decisions are made based on converging evidence from disparate sources.

\bibliography{MTL_AuxLabelInfo}
\end{document}


\title{Multi-task Learning with Weak Class Labels: Leveraging iEEG to Detect Cortical Lesions in Cryptogenic Epilepsy}

\author{\name Supplementary Material \\
       } 

\maketitle

\section{MTL with Rank Constraints}
Here we provide the primal problems for both the globally consistent and task-specific label ranking appproaches. We also provide the details of the dual optimization problem for both approaches.

\subsection{MTL with globally-consistent label ranking:} 
The primal optimization problem is given as:
\small
\begin{equation}
\label{eqn:primalMTLwithConstraints}
\begin{aligned}
&& \underset{w_{0},v_{t},\xi_{i}^{t},\eta_{pq}^{t}}{\text{min}}
& \left\{ \frac{1}{2} \sum_{t=1}^{T} {\left\| v_{t} \right\|}^{2} + \frac{\mu}{2} {\left\| w_{0} \right\|}^{2} + \right.\\
&&& \left. + C \sum_{t=1}^{T}\sum_{i=1}^{m}\xi_{i}^{t} + C' \sum_{t=1}^{T}\sum_{\left(p,q\right) \in \Pi_{t}}\eta_{pq}^{t} \right\} \\
&&\text{subject to:} & \\
&&\forall i , \ \ \forall t:\ &\ y_{i}^{t} \left(w_{0}+v_{t}\right) \cdot x_{i}^{t} \  \geq \ 1-\xi_{i}^{t} \\ 
&&\forall t, \forall \left(p,q\right) \in \Pi_{t}:\ &z_{pq}^{t}\left(w_{0}+v_{t}\right) \cdot \Delta_{pq}^{t} \ \geq \ 1-\eta_{pq}^{t}\\
&& \ & \ \xi_{i}^{t} \ \ \geq \ 0 \ \ ,\ \  \ \eta_{pq}^{t} \ \ \geq \ 0
\end{aligned}
\end{equation}
\normalsize
where $\eta_{pq}^{t}$ and $\xi_{i}^{t}$ are slack variables and $z_{pq}^{t}=1$ are the labels for each pseudo-example: $\Delta_{pq}^{t}=x_{p}^{t}-x_{q}^{t}$. Note that, pseudo-examples are created on a per task basis i.e., there are no pseudo-examples resulting from comparing the ranks of two instances from different tasks. The dual of above problem can be formulated as:
\small
\begin{equation}
\label{eqn:mainMTLProblemDualWithConstraints}
\begin{aligned}
& \underset{\alpha_{i}^{t},\beta_{pq}^{t}}{\text{max}} \  
 \left\{\sum_{t=1}^{T}\sum_{i=1}^{m}\alpha_{i}^{t} +  \sum_{t=1}^{T}\sum_{(p,q) \in \Pi_{t}}\beta_{pq}^{t}    \right. - \\
& \left. \frac{1}{2} \sum_{t=1}^{T}\sum_{i=1}^{m}\sum_{s=1}^{T}\sum_{j=1}^{m}{\alpha_{i}^{t}y_{i}^{t}\alpha_{j}^{s}y_{j}^{s}K_{st}\left(x_{i}^{t},x_{j}^{s}\right)} \right. - \\
& \left. - \sum_{t=1}^{T}\sum_{i=1}^{m}\sum_{s=1}^{T}\sum_{(p,q) \in \Pi_{t}}{\alpha_{i}^{t}y_{i}^{t}\beta_{pq}^{s}z_{pq}^{s}K_{st}\left(x_{i}^{t},\Delta_{pq}^{s}\right)} \right. - \\
& \left.  - \frac{1}{2} \sum_{t=1}^{T}\sum_{(p,q) \in \Pi_{t}}\sum_{s=1}^{T}\sum_{(k,l) \in \Pi_{s}}{\beta_{pq}^{t}z_{pq}^{t}\beta_{kl}^{s}z_{kl}^{s}K_{st}\left(\Delta_{pq}^{t},\Delta_{kl}^{s}\right)} \right\} \\
& \text{subject to:} \\
& \ \ \ \forall i , \ \ \forall t:\   0 \leq \alpha_{i}^{t} \leq C \ , \  0 \leq \beta_{pq}^{t} \leq C'
\end{aligned}
\end{equation}
\normalsize
where, $\alpha_{i}^{t}$ and $\beta_{pq}^{t}$ are Lagrange multipliers corresponding to the classification and rank constraints, respectively. $K_{ij}(.,.)$ is defined as  \citep{Evgeniou:2004:RegMultiTask}:
\small
\begin{equation}
\label{eqn:MTLKernelForm}
K_{st}(x_{i}^{t},x_{j}^{s}) = \left( \frac{1}{\mu} + \delta_{st} \right) x_{i}^{t} \cdot x_{j}^{s}
\end{equation}
\normalsize

The optimal solutions for both $w_{0}$ and $v_{t}$ can be obtained by formulating and solving the Lagrangian function for Equation \ref{eqn:primalMTLwithConstraints}.
\small
\begin{equation}
\label{eqn:w0GlobalConsitency}
w^{*}_{0}=\frac{1}{\mu}\left[ \sum_{t=1}^{T}\sum_{i=1}^{m}\alpha_{i}^{t}y_{i}^{t}x_{i}^{t}+\sum_{s=1}^{T}\sum_{{\left(p,q\right) \in \Pi_{s}}}\beta_{pq}^{s}z_{pq}^{s}\Delta_{pq}^{s} \right]
\end{equation}
\begin{equation}
\label{eqn:vTGlobalConsitency}
v^{*}_{t}= \sum_{i=1}^{m}\alpha_{i}^{t}y_{i}^{t}x_{i}^{t}+\sum_{{\left(p,q\right) \in \Pi_{t}}}\beta_{pq}^{t}z_{pq}^{t}\Delta_{pq}^{t} 
\end{equation}
\normalsize

\subsection{MTL with task-specific label ranking:} 
In this formulation we limit the influence of rank constraints on the overall solution, by restricting them to affect only their own task-specific components. This is formulated as:
\small
\begin{equation}
\label{eqn:primalMTLwithLocalConstraints}
\begin{aligned}
&& \underset{w_{0},v_{t},\xi_{i}^{t},\eta_{pq}^{t}}{\text{min}}
& \left\{ \frac{1}{2} \sum_{t=1}^{T} {\left\| v_{t} \right\|}^{2} + \frac{\mu}{2} {\left\| w_{0} \right\|}^{2} + \right.\\
&&& \left. C \sum_{t=1}^{T}\sum_{i=1}^{m}\xi_{i}^{t} + C' \sum_{t=1}^{T}\sum_{\left(p,q\right) \in \Pi_{t}}\eta_{pq}^{t} \right\} \\
&&\text{subject to:} & \\
&&\forall i , \ \ \forall t:\ &\ y_{i}^{t} \left(w_{0}+v_{t}\right) \cdot x_{i}^{t} \  \geq \ 1-\xi_{i}^{t} \\ 
&&\forall t, \forall \left(p,q\right) \in \Pi_{t}:\ &\ z_{pq}^{t} \  v_{t} \cdot \Delta_{pq}^{t}\geq 1-\eta_{pq}^{t}\\
&& \ & \ \xi_{i}^{t} \ \ \geq \ 0 \ \ ,\ \  \ \eta_{pq}^{t} \ \ \geq \ 0
\end{aligned}
\end{equation}
\normalsize
By not allowing the rank information to directly influence the shared component $w_{0}$, the ranking function $\Pi_{t}$ is no longer coupled across tasks, and can behave differently for different tasks. It should be noted that although the shared weight vector $w_{0}$ is not required to preserve the rankings, it is still indirectly affected by the rank constraints through $v_{t}$ (since, $w_{t}=w_{0}+v_{t}$). The dual for Equation \ref{eqn:primalMTLwithLocalConstraints} can be formulated in a form identical to Equation \ref{eqn:mainMTLProblemDualWithConstraints}, by using a new kernel function. Let $\tilde{X}^{t} \in \mathbb{R}^{d}$ be the augmented data for task $t$ obtained by combining all the original data instances ($x_{i}^{t}$) and the pseudo-examples ($\Delta_{pq}^{t}$), and let $u_{i}^{t}$ be an indicator variable defined as:
\small
\begin{equation*}
u_{k}^{t}= 
\begin{cases} 
1 & \text{if $\tilde{x}_{k}^{t}=x_{i}^{t}$,}
\\
0 &\text{if $\tilde{x}_{k}^{t}=\Delta_{pq}^{t}$.}
\end{cases}
\end{equation*}
\normalsize
where, $k \in \{1,\dots,|\tilde{X}^{t}|\}$. Let the kernel function be defined as:
\small
\begin{equation}
\label{eqn:MTLKernelNew}
K_{st}(\tilde{x}_{k}^{t},\tilde{x}_{l}^{s}) = \left( \frac{u_{k}^{t} u_{l}^{s}}{\mu} + \delta_{st} \right) \tilde{x}_{k}^{t} \cdot \tilde{x}_{l}^{s}
\end{equation}
\normalsize
This is an operator-valued kernel \citep{Evgeniou:2005:LMT} that does not allow the ranking function $\Pi_{t}$ to directly impact $w_{0}$, thus, restricting the auxiliary label information from being directly shared among tasks. 

In this formulation, the optimal solution for $v_{t}$ does not change and remains identical to Equation \ref{eqn:vTGlobalConsitency}. The difference lies in the optimal solution for the shared parameter vector $w_{0}$, which in this case is given as:
\small
\begin{equation}
\label{eqn:w0TaskSpecific}
w^{*}_{0}=\frac{1}{\mu} \sum_{t=1}^{T}\sum_{i=1}^{m}\alpha_{i}^{t}y_{i}^{t}x_{i}^{t}
\end{equation}
\normalsize
As expected, it can be seen that $w_{0}$ is no longer affected by the rank constraints.

\section{Imaging}
Participants were selected from a large registry of patients with epilepsy. Criteria for inclusion in this study included: (1) completion of a high resolution T1-weighted MRI scan; (2) surgical resection to treat focal epilepsy; (3) diagnosis of FCD on neuropathological examination of the resected tissue. 

\subsection{Imaging Protocol} 
Imaging was performed on a Siemens Allegra 3T scanner. Image acquisitions included a conventional 3-plane localizer and a T1-weighted volume pulse sequence (TE=3.25 ms, TR =2530 ms, TI =1100 ms, flip angle =7 deg field of view (FOV) = 256 mm, matrix = 256x256, vertex size =1x1x1.3 mm, scan time: 8:07 min). Acquisition parameters were optimized for increased gray/white matter image contrast. The T1-weighted image was reoriented into a common space, roughly similar to alignment based on the AC-PC line. Images were corrected for nonlinear warping caused by no-uniform fields created by the gradient coils. 

\subsection{Surface Extraction}
The MRI sequences were processed using the FreeSurfer software package\footnote{Available at http://surfer.nmr.mgh.harvard.edu/}, which performs automated tissue segmentation to recreate 3D representations of the cortical surfaces from structural MRI scans \citep{Dale:SBA1:1999}. Briefly, after skull striping, the method \citep{Dale:SBA1:1999} involves: (i) segmentation of the white matter, (ii) tessellation of the gray/white matter boundary, (iii) inflation of the folded surface, and (iv)correction of topological defects. Once the surface was reconstructed it was further refined by classifying all white matter vertices in the MRI volume to create the gray/white matter boundary. The gray/white matter junction was delineated up to submillimeter accuracy by further refining the white matter surface. After refining the gray/white matter junction the pial surface was located by deforming the surface outward. Each segmentation and reconstruction underwent manual inspection and editing, when necessary. However, the high image quality and gray-white contrast in the initial images resulted in minimal editing requirements for both patient and control scans. Surface reconstruction was followed by a registration process that involved morphing the reconstructed surface to an average spherical representation that accurately matched sulcal and gyral features across individual subjects while minimizing metric distortion \citep{Fischl:SBA2:1999}.

\section{Model Hyperparameters}
In this section we explain the different model hyperparamaters for the proposed methods and the baselines. We also provide the details of how their values were set in our experiments. All model parameters for the proposed approaches and the baselines were set using the data of three MRI-negative patients and their matched controls. No iEEG data was available for these three patients, and hence we used them for setting the model hyperparameters. We will refer to this set of three patients as the model parameter set (MPS). The data for these three patients are distinct from the fifteen patients and controls used in our experiments and whose results are reported in the paper.

\subsection{Segmentation} 
The standard quick shift algorithm is a fast mode seeking algorithm similar to mean shift \citep{Vedaldi:quickshift:2008}. It performs a hierarchical segmentation of the image, where the sub-trees represent image segments. It has two parameters namely the size of the Gaussian kernel ($\sigma_{QS}$) used by a Parzen window density estimator, and the maximum distance ($\delta_{QS}$) between two pixels permitted while remaining part of the same segment. The scale parameter $\sigma_{QS}$ is varied to change the average size of segments, and $\delta_{QS}$ is set to be a multiple of $\sigma_{QS}$ \citep{Vedaldi:quickshift:2008}. Thus, higher values of $\sigma_{QS}$ produce larger segments. All the patients and controls were segmented using the same set of Quickshift parameters ($\sigma_{QS}=8, \delta_{QS}=32$) that were optimized using the MPS.

\begin{table}[t]
\renewcommand{\arraystretch}{1.3}
\centering
\begin{tabular}{c|c}
\hline
{Parameter} & {Range} \\
\hline
{$\mu$} & {$10^{-7},5^{-6},10^{-6},5^{-5},\ldots,10^{3}$} \Tstrut \\
{$C$} & {$2^{-10},2^{-9},\ldots,2^{10}$} \\
{$\gamma$} & {$2^{-10},2^{-9},\ldots,2^{10}$}\\
{$a$} & {$10^{-6},5^{-5},10^{-5},\ldots,10^{3}$}\\
\hline
\end{tabular}
\caption{Range of values for the model hyper-parameters used in the grid search. The grid search optimized the area under the curve (AUC) over the model parameter set (MPS) consisting of three patients whose data is distinct from the fifteen patients used for performance analysis.}
\label{table:GridDesign}
\end{table}

\subsection{Setting the Hyper-Parameters}
It should be noted that the hyper-parameters were set individually for each test subject using the MPS. Below we provide the details about how the parameters were set for the different baselines and our proposed methods:
\begin{itemize}
\item \emph{LDA:} The detection threshold ($\tau \in \left[0,1\right]$) for LDA was optimized by maximizing the area under the curve (AUC) over the MPS.
\item \emph{SLR:} The detection threshold ($\tau \in \left[0,1\right]$) for logistic regression was optimized by maximizing the area under the curve (AUC) over the MPS.
\item \emph{SVMR:} We also used a single-task SVM using the RBF kernel that incorporates ranking constraints based on the model in \citep{Nguyen:2011:LearningAuxProb} as a baseline. In addition to the cost parameter (identical to the cost parameter of tradtional SVM) and the scale parameter of the RBF kernel, there is a third parameter $a$ that defines the relative cost of violating a ranking constraint ($C'=aC$). All three parameters were set by optimizing the AUC using the MPS.
\item \emph{MTL:} This corresponds to the regularized MTL framework that does not incorporate any auxiliary supervision, and uses the resection zone as class labels. The model parameters were set using a grid-search \citep{Evgeniou:2004:RegMultiTask} and include mis-classification cost ($C$), task-relatedness parameter ($\mu$) and the scale ($\gamma$) of the RBF kernel. To find suitable values for the parameters we used a three-level grid and optimized the area under the curve (AUC) over the MPS. 
\item \emph{GC \& TS:} These are the proposed methods that incorporate auxiliary supervision derived from iEEG data. In addition to the three model parameters for MTL, there is a fourth parameter $a$ that defines the relative cost $C'=aC$ of violating a rank constraint (c.f., Equations \ref{eqn:primalMTLwithConstraints} and \ref{eqn:primalMTLwithLocalConstraints}). To find suitable values for the parameters we designed a four-level grid and optimized the AUC over the MPS. 
\end{itemize}
Table 1 lists the ranges for the parameters used in the grid search for SVMR, MTL, GC and TS.

\bibliography{MTL_AuxLabelInfo}